\documentclass{article}
\usepackage{ijcai22}

\usepackage{times}
\usepackage{soul}
\usepackage{url}
\usepackage[hidelinks]{hyperref}
\usepackage[utf8]{inputenc}
\usepackage[small]{caption}
\usepackage{graphicx}
\usepackage{amsmath}
\usepackage{amsthm}
\usepackage{booktabs}
\usepackage{algorithm}
\usepackage{algorithmic}
\usepackage{microtype}

\usepackage{amsmath} 
\usepackage{amssymb}            
\usepackage{amsfonts}           
\usepackage{mathrsfs}   
\usepackage{multirow}           
\usepackage{comment}
\usepackage{graphicx} 
\usepackage{bm}
\usepackage{booktabs}
\usepackage{makecell}

\title{Generating a Structured Summary of Numerous Academic Papers: \\Dataset and Method}

\author{
Shuaiqi LIU
\and
Jiannong Cao\and
Ruosong Yang\And
Zhiyuan Wen
\affiliations
Department of Computing, The Hong Kong Polytechnic University\\
\emails
\{cssqliu, csjcao, csryang, cszwen\}@comp.polyu.edu.hk
}

\begin{document}

\maketitle

\begin{abstract}
Writing a survey paper on one research topic usually needs to cover the salient content from numerous related papers, which can be modeled as a multi-document summarization (MDS) task. Existing MDS datasets usually focus on producing the structureless summary covering a few input documents. 
Meanwhile, previous structured summary generation works focus on summarizing a single document into a multi-section summary. 
These existing datasets and methods cannot meet the requirements of summarizing numerous academic papers into a structured summary. To deal with the scarcity of available data, we propose BigSurvey, the first large-scale dataset for generating comprehensive summaries of numerous academic papers on each topic. We collect target summaries from more than seven thousand survey papers and utilize their 430 thousand reference papers’ abstracts as input documents. To organize the diverse content from dozens of input documents and ensure the efficiency of processing long text sequences, we propose a summarization method named category-based alignment and sparse transformer (CAST). The experimental results show that our CAST method outperforms various advanced summarization methods. 
\end{abstract}

\section{Introduction}

The number of published academic papers has been growing rapidly \cite{aviv2021publication}. 
It brings difficulties for researchers to read through the numerous papers on the research topics they are interested in.
A summary of papers on a research topic can help researchers quickly browse key information in these papers.
As a type of human-written summary, the survey paper can review numerous papers on each research topic and guide people to learn the topic. But writing a survey paper needs a lot of time and effort, making it difficult to cover the latest papers and all the research topics. 
The multi-document summarization (MDS) techniques \cite{liu2018generating,fabbri2019multi,liu2021highlight,LIU2022102913} can be utilized to automatically produce summaries as a supplement to human-written summaries.
To cover the latest papers and more research topics at a low cost, people can flexibly adjust the input papers and let the summarization methods produce summaries for these papers.
Our target is to generate the comprehensive, well-organized, and non-redundant summary for numerous papers on the same research topic. 
To achieve this target, there are some challenging issues, including the scarcity of available data, the organization of diverse content from different sources, and summarization models' efficiency in processing long texts. 

Although there have been some MDS datasets \cite{fabbri2019multi,lu-etal-2020-multi-xscience}, most of them focus on producing short and structureless summaries covering less than ten input documents, which cannot meet the real needs of reviewing numerous papers on one research topic.
To deal with the scarcity of available data, we propose BigSurvey, the first large-scale dataset for numerous academic papers summarization.
It contains more than seven thousand survey papers and their 434 thousand reference papers' abstracts. 
Considering copyright issues, we collect these reference papers' abstracts as input documents for MDS. These abstracts can be regarded as summaries written by their authors, which include these reference papers' salient information.

These input abstracts usually have content on multiple aspects, including the background, method, objective, and results. It is challenging for a summary to organize and present the diverse content from dozens of input documents. Compared with the structureless summary, the structured summary contains multiple sections summarizing particular aspects of input content and is found easier to read and more welcomed by readers \cite{hartley2004current,hartley2014current}.
To balance the comprehensiveness and brevity, we built two subsets of the BigSurvey for producing two-level summaries. 
The BigSurvey-MDS focuses on producing comprehensive summaries, while the BigSurvey-Abs is built for producing more concise summaries of these summaries in BigSurvey-MDS.

\begin{figure*}[t]
\centering
\includegraphics[width=6.6in]{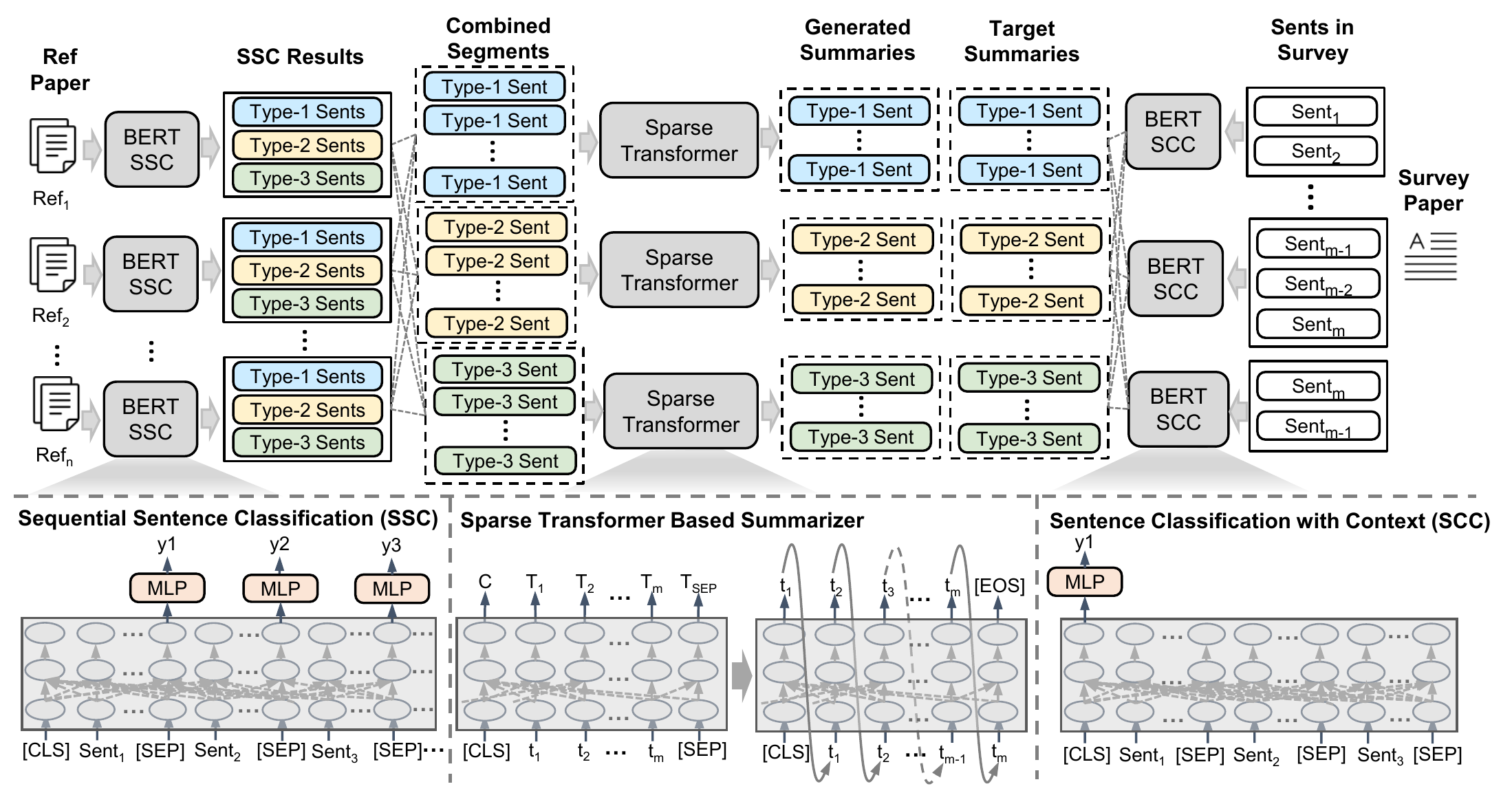}
\caption{An overview of our CAST method.}\label{fig:diversity_visual} 
\end{figure*} 

We make two assumptions for the structured summary of multiple papers on the same topic.
1) the research topic's descriptions on one aspect is a subset of the union of related papers' content on this aspect (e.g., the research topic's background should be part of all the related papers' background). 
2) Each section of the structured summary focuses more on the salient content in one subset mentioned in 1) (e.g., the summary's background section focuses on the salient content in all the reference papers' background).
Based on these assumptions, we propose the category-based alignment (CA) to align each section of the structured summary with a set of input sentences classified as the same type.

As shown in Table \ref{table:stats_dataset_all}, the average sum of input documents' word number is close to twelve thousand in each example of the BigSurvey dataset. 
The much longer inputs can introduce more noises, and the salient content can be more scattered, which makes it more difficult to capture and encode the salient content. Long input sequences can also reduce the efficiency of summarization models since existing neural models' time or space complexity is usually highly correlated with the input sequence length.
To deal with the above problems, we propose a method named category-based alignment and sparse transformer (CAST). As shown in Figure \ref{fig:diversity_visual}, we use the BERT-based sequential sentence classification (SSC) method and the sentence classification with context (SCC) method to classify input and output sentences. Then, we use the category-based alignment to align the sets of input and target output sentences classified as the same type and compose examples for training summarization models.
We adopt the transformer with the sparse attention mechanism for abstractive summarization. The sparse attention supports the encoder to model longer input sequences with limited GPU memory. 
Our BigSurvey dataset and CAST method make it possible to finetune a large pre-trained model to generate structured summaries covering dozens of input documents on an off-the-shelf GPU.

We benchmark advanced extractive and abstractive summarization methods as baselines on our BigSurvey dataset. To compare their performance, we conduct automatic evaluation and human evaluation. Experimental results show that our proposed CAST method outperforms these baseline models, and adding the category-based alignment can bring extra performance gains for various summarization methods.

Our contribution is threefold:
\begin{itemize}
\item We build BigSurvey, the first large-scale dataset for numerous academic papers summarization.
\item We propose a method named category-based alignment and sparse transformer (CAST) to summarize numerous academic papers on each research topic. 
\item We benchmark various summarization methods on our dataset and find that adding the category-based alignment can bring extra performance gains for various methods.
\end{itemize}

\section{Related Work}
Some large-scale MDS datasets \cite{fabbri2019multi,liu2018generating,lu-etal-2020-multi-xscience} have been released in these years, which makes it possible to train large neural models for MDS. Some of these datasets are relevant to our work.
Multi-XScience \cite{lu-etal-2020-multi-xscience} is a scientific paper summarization dataset. Its target summaries are individual paragraphs in scientific papers' related work sections. Each of these summaries has less than five input documents on average.
Another dataset WikiSum \cite{liu2018generating} aims to generate the first section of the Wikipedia article. Since most of these articles have a few references, they supplement the input with more search results. 
Unlike these existing MDS datasets, our BigSurvey dataset is for producing comprehensive summaries to cover numerous academic papers on each research topic.

\begin{table*}[t]
\small
\renewcommand\arraystretch{1.0}
\centering
\begin{tabular}{lccccccccc}
\toprule \textbf{Dataset} & \textbf{Pairs} & \textbf{\makecell*[c]{Words\\ (Doc)}} & \textbf{\makecell*[c]{Sents\\ (Doc)}}& \textbf{\makecell*[c]{Words\\ (Sum)}} & \textbf{\makecell*[c]{Sents\\ (Sum)}} & \textbf{\makecell*[c]{Input Doc\\ Num}} & \textbf{Cov.} & \textbf{Dens.} &\textbf{Comp.} \\
\midrule
Multi-News & 56,216 & 2,103.5 & 82.7 & 263.7 & 10.0 & 2.8 & 0.69 & 3.1 & 6.3\\
Multi-XScience & 40,528 & 778.1 & 23.7 & 116.4 & 4.9 & 4.4 & 0.60 & 1.1 & 5.6\\
PubMed & 133,215 & 3,049.0 & 87.5 & 202.4 & 6.8 & 1 & 0.79 & 4.3 & 13.6\\ 
ArXiv & 215,913 & 6,029.9 & 205.7 & 272.7 & 9.6 & 1 & 0.87 & 3.8 & 39.8\\
\midrule
BigSurvey-MDS & 4,478 & 11,893.1 & 450.1 & 1,051.7 & 38.8 & \textbf{76.3} & 0.81 & 1.5 & 11.3\\
BigSurvey-Abs & 7,123 & 12,174.5 & 463.8 & 170.1 & 6.4 & 1 & 0.83 & 3.5 & 71.6\\
\bottomrule
\end{tabular}
\caption{Comparison of our BigSurvey dataset to other summarization datasets. "Pairs" denotes the number of examples. "Words" and "Sents" indicate the average number of words and sentences in input text or target summary. "Input Doc Num" represents the average number of input documents in each example. "Cov." is the extractive fragment coverage, "Dens." is the extractive fragment density, and "Comp." is the compression ratio of target summaries.}
\label{table:stats_dataset_all}
\end{table*}

As for producing the structured summary, there have been many single document summarization (SDS) works. 
Gidiotis and Tsoumakas~\shortcite{GidiotisT19} build the PMC-SA dataset, in which target summaries are papers' structured abstracts containing multiple sections (e.g., the introduction, methods, results, and discussion). To compose the pairs of input and output, they match body sections with abstract's sections by section titles. 
Meng et al.~\shortcite{meng2021bringing} also align the input document's sections with target summary's sections.
These SDS works can utilize the explicit formats (e.g., the division of sections) of input documents and target summaries to determine the alignment relationships between the input and output.
For multi-document summarization, input documents can be collected from different sources (journals or conferences) and follow diverse formats. Our collected reference papers' abstracts usually do not have sections, so we can not use the section-level alignment on our dataset. Finding alignment relationships between input and output content becomes a challenge.

\section{BigSurvey Dataset}
\label{sec:dataset}
In this section, we first present our data sources and procedures of data collection and pre-processing. And then, we introduce our BigSurvey dataset\footnote{Our dataset: https://github.com/StevenLau6/BigSurvey}.
We also conduct the descriptive statistics and in-depth analysis of our dataset and compare them with other commonly used document summarization datasets. 

\subsection{Data Collection and Pre-processing}
\label{subsec:data_collect_preprocess}

We collect more than seven thousand survey papers from arXiv.org \footnote{These survey papers' metadata are collected from a June 2021 dump (https://www.kaggle.com/datasets/Cornell-University/arxiv)}, download their PDF files by their dois, and parse these files with a tool named science-parse\footnote{https://github.com/allenai/science-parse}. 
We can extract the bibliography information (e.g., reference papers' titles and authors) from parsing results. 
Based on these survey papers' bibliography information, we collect their reference papers' abstracts from Microsoft Academic Service (MAS) \cite{sinha2015overview} and Semantic Scholar \cite{ammar2018construction}.
We collected more than 434 thousand reference papers in total.

In the pre-processing stage, we first filter out invalid samples from collected data. Specifically, downloaded files that are duplicated or could not be parsed properly (e.g., some PDF files are scanned or incomplete) are removed. We also filter out outliers with too-short parsed texts in the survey papers or very few collected reference papers. 
For each selected survey paper, we remove noises (e.g., the copyright information before the first section and special symbols used to compose a style), extract the abstract and introduction section from these survey papers, and truncate their reference papers' abstracts.
We lowercase these texts and use NLTK \cite{bird2009natural} to split sentences and words.
After that, we split the training (80\%), validation (10\%), and test (10\%) sets.

\subsection{Dataset Description}
BigSurvey is a large-scale dataset containing two-level target summaries for dozens of academic papers on the same topic. The long summary aims to comprehensively cover the reference papers' salient content in different aspects, while the much shorter summary is more concise and can be regarded as the summary of the long summary.
For these two-level summaries, we build two subsets: BigSurvey-MDS and BigSurvey-Abs. Their statistical information is shown in Table \ref{table:stats_dataset_all}. We will introduce their definitions and properties separately.

\paragraph{BigSurvey-MDS.} 
This subset focuses on producing comprehensive summaries covering numerous academic papers on one research topic.
Each example in the BigSurvey-MDS corresponds to one survey paper from arXiv.org. These survey papers usually have tens or hundreds of reference papers.
Considering copyright issues, BigSurvey-MDS does not include these reference papers' body sections and uses their abstracts as input documents. 
These abstracts can be regarded as summaries written by their authors, which include these papers' salient information.
For each survey paper, we collect at most two hundred reference papers' abstracts and truncate each of them to no more than two hundred words. These truncated abstracts are used as input documents of the BigSurvey-MDS.

The survey paper's introduction section usually introduces a research topic's background, method, and other aspects.
We split the content of the survey paper's introduction into three sections (the background, method, and other) and use them to compose the structured summary as the target in each example of the BigSurvey-MDS.
The content about the objective, result, and other are merged into the section named other because we observe that these types of content appear less frequently than the background and method in the survey papers' introduction section. 
To prepare these three sections in the target summary, we first collect the introduction section from a survey paper. If there is no introduction section, we extract the survey paper's first 1,024 words after the abstract part.
Then we classify sentences in the introduction section and concatenate the sentences classified as the same type to form the three sections in the target summary. 
We filter out the examples with too short input sequences or target summaries. 
As shown in Table \ref{table:stats_dataset_all}, BigSurvey-MDS's average input length, average output length, and the average number of input documents are much larger than previous MDS datasets.

\begin{table}[t]
\small
\renewcommand\arraystretch{1.0}
\centering
\setlength{\tabcolsep}{1.2mm}{
\begin{tabular}{lcccc}
\toprule \textbf{\multirow{2}*{Dataset}} & \multicolumn{4}{c}{\textbf{\% of novel n-grams in target summary}} \\
~ & \textbf{unigrams} & \textbf{bigrams} & \textbf{trigrams} & \textbf{4-grams} \\
\midrule
Multi-News & 17.76 & 57.10 & 75.71 & 82.30\\
Multi-XScience & 42.33 & 81.75 & 94.57 & 97.62\\
PubMed & 18.38 & 49.97 & 69.21 & 78.42\\
ArXiv & 15.04 & 48.21 & 71.66 & 83.26\\

\midrule
BigSurvey-MDS & 37.39 & 76.46 & 93.87 & 98.04\\
BigSurvey-Abs & 19.85 & 53.97 & 74.15 & 82.22\\

\bottomrule
\end{tabular}}
\caption{The proportion of novel n-grams in target summaries.}
\label{table:stats_dataset_novel_ngrams}
\end{table}

\begin{figure}[t]
\centering
\includegraphics[width=3.2in]{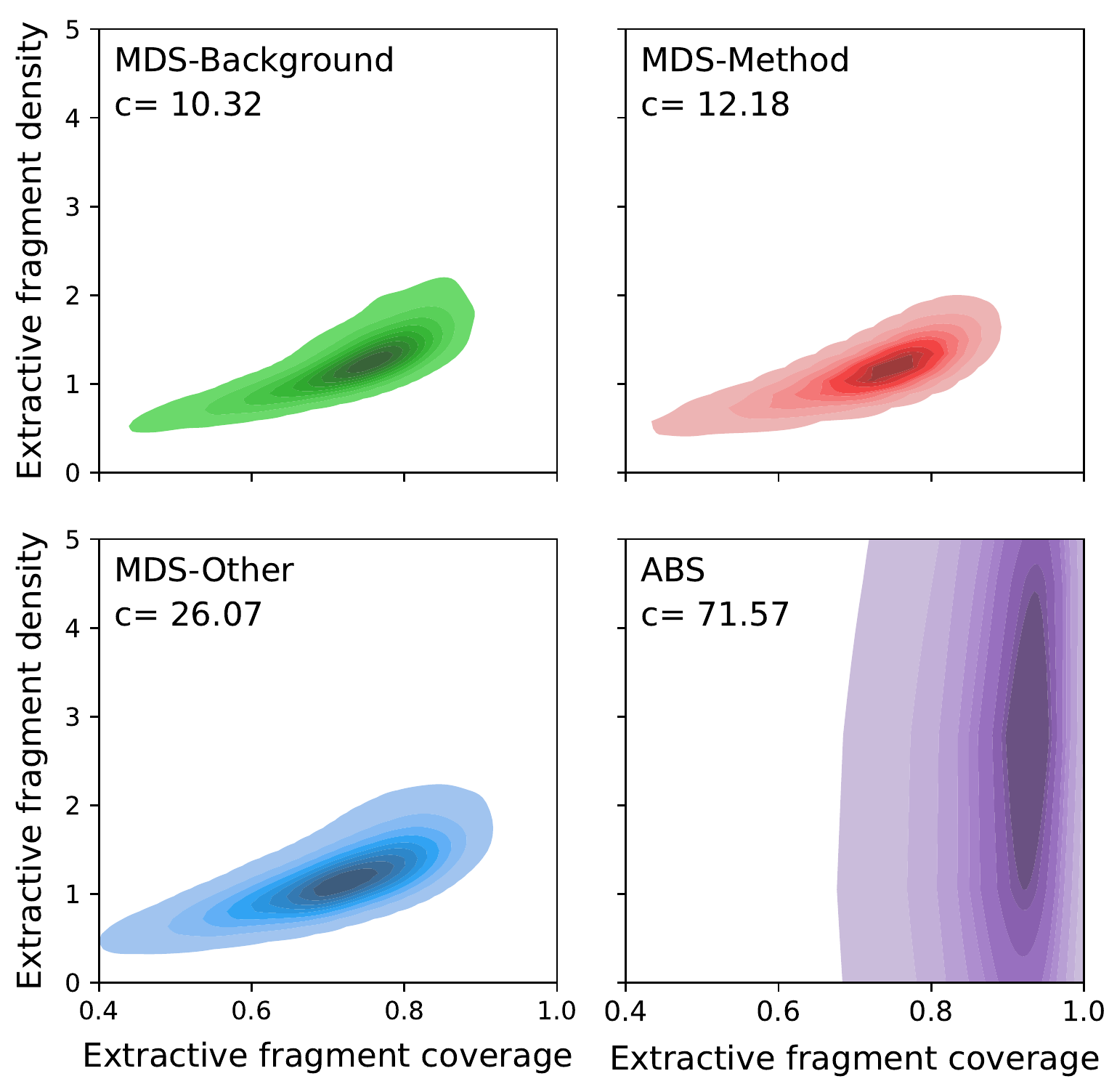}
\caption{Coverage and density distributions of the BigSurvey.}\label{fig:diversity_visual} 
\end{figure}

\paragraph{BigSurvey-Abs.} The body text of a survey paper can be regarded as a comprehensive and long summary of its reference papers. Meanwhile, the survey paper's abstract is a short summary of its body text. 
The subset named BigSurvey-Abs uses these survey papers' abstracts as target summaries, which aims to produce more concise summaries of these survey papers' body text.
Considering the constraints of GPU memory, we truncate these survey papers in our experiments. Specifically, we follow the settings in \cite{zhang2020pegasus,zaheer2020big} to use the first 1,024 words as the input for transformer-based models without sparse attention and use the first 3,072 words as the input for transformer-based models with sparse attention. 
In this case, the input documents of the BigSurvey-Abs highly overlap with the target summaries in the BigSurvey-MDS. Therefore, the short summary in the BigSurvey-Abs can be regarded as the summary of the long summary in the BigSurvey-MDS.
Besides, the average input and output lengths are similar to previous academic literature summarization datasets. Previous text summarization methods should be able to adapt to the BigSurvey-Abs dataset.

\subsection{Diversity Analysis of Dataset}
\label{subsec:appendix_diversity}

To measure how abstractive our target summaries are, we report the percentage of target summaries' novel n-grams, which do not appear in input documents. Table \ref{table:stats_dataset_novel_ngrams} reflects that the abstractiveness of the BigSurvey-MDS subset is similar to that of the Multi-XScience. The BigSurvey-Abs subset's abstractiveness is lower than that of the BigSurvey-MDS and Multi-XScience, and it is similar to other existing datasets.

Besides, we assess the extractive nature of the BigSurvey's subsets by using three measures defined by Grusky et al.~\shortcite{grusky-etal-2018-newsroom}, including the extractive fragment coverage, extractive fragment density, and compression ratio. The extractive fragment coverage measures the percentage of words in the summary that are part of an extractive fragment from the input document. The extractive fragment density assesses the average length of the extractive fragment where each word in the target summary belongs. The compression ratio is the word ratio between the input documents and their target summaries.
Results of these three measures are visualized using kernel density
estimation. 
Figure \ref{fig:diversity_visual} shows that three summary sections in the BigSurvey-MDS subset have similar distributions in coverage and density. Their densities are low, and their coverages vary in a relatively large range. The BigSurvey-Abs subset varies largely along the y-axis (extractive fragment density), which suggests varying writing styles of target summaries.

\section{Method}
\label{sec:method}
In this paper, we propose a solution named category-based alignment and sparse transformer (CAST) to summarize numerous academic papers on one research topic. CAST contains three main components: the BERT-based sentence classification with context (SCC) model, the sequential sentence classification (SSC) model, and the transformer-based abstractive summarization model with sparse attention.

Each section of the structured summary usually focuses on a specific aspect of the content from input documents.
To prepare each summary section's content, we classify sentences in the extracted introduction section of a survey paper and merge sentences classified as the same type.
We design a method named sentence classification with context (SCC) to classify these sentences. Given a sentence and the sentences before and after it, we concatenate them as the input for the sentence classification model based on a pre-trained model (e.g., BERT \cite{devlin2019bert} or RoBERTa \cite{liu2019roberta}). We train the SCC model on the labeled sentences from the CSABST dataset \cite{cohan2019pretrained}, in which each sentence is annotated as one of 5 categories: background, objective, method, result, and other.

\begin{table*}[t]
\renewcommand\arraystretch{1.1}
\small
\centering
\setlength{\tabcolsep}{1.9mm}{
\begin{tabular}{lccccccccccccccc}
\toprule \textbf{\multirow{2}*{Method}} & \multicolumn{3}{c}{\textbf{Background}} & &\multicolumn{3}{c}{\textbf{Method}} & & \multicolumn{3}{c}{\textbf{Other}} & & \multicolumn{3}{c}{\textbf{Combined}}\\
\cline{2-4} \cline{6-8} \cline{10-12} \cline{14-16}
~ & \textbf{R-1} & \textbf{R-2} & \textbf{R-L} & &\textbf{R-1} & \textbf{R-2} & \textbf{R-L} & & \textbf{R-1} & \textbf{R-2} & \textbf{R-L}& & \textbf{R-1} & \textbf{R-2} & \textbf{R-L}\\
\hline
LexRank & - & - & - & & - & - & - & & - & - & - & & 35.85 & 8.59 & 14.22\\
LexRank+CA & 31.33 & 5.92 & 13.93 & & 28.85 & 4.65 & 13.07 & & 23.61 & 6.08 & 13.04 & & 37.92 & 8.56 & 14.63\\
TextRank & - & - & - & & - & - & - & & - & - & - & & 36.35 & 8.49 & 14.24\\
TextRank+CA & 31.20 & 5.79 & 13.91 & & 28.80 & 4.38 & 12.94 & & 24.41 & 6.42 & 13.81 & & 38.22 & 8.45 & 14.70\\
\hline
BART & 31.96 & 5.73 & 14.96 & & 28.61 & 4.97 & 14.32 & & 23.87 & 5.74 & 13.50 & & 37.64 & 8.45 & 15.69\\
BART+CA & 33.05 & 6.21 & 15.40 & & 29.22 & 5.22 & 14.57 & & 25.39 & 6.58 & 14.44 & & 40.21 & 9.38 & 16.06 \\
PEGASUS & 33.51 & 6.74 & 15.67 & & 27.47 & 4.93 & 14.17 & & 25.20 & 6.55 & 14.02 & & 38.91 & 9.00 & 16.20 \\
PEGASUS+CA & 33.93 & 6.80 & 15.67 & & 29.76 & 5.74 & 15.05 & & 26.32 & 7.34 & 15.34 & & 41.09 & 9.96 & 16.76\\
BigBird-PEGASUS & 34.31 & 6.78 & 15.54 & & 29.46 & 5.47 & 14.43 & & 26.07 & 6.66 & 14.24 & & 41.29 & 9.84 & 16.37\\
LED & 34.11 & 6.84 & 15.78 & & 26.15 & 4.59 & 13.47 & & 25.26 & 6.34 & 13.74 & & 39.79 & 9.42 & 16.05 \\
CAST-BigBird & 34.56 & 6.96 & 15.55 & & 30.83 & 5.95 & 15.03 & & 26.90 & 7.47 & 15.45 & & 42.10 & 10.24 & 16.71\\
CAST-LED & \textbf{36.55} & \textbf{8.82} & \textbf{16.87} & & \textbf{31.72} & \textbf{6.94} & \textbf{15.61} & & \textbf{27.16} & \textbf{8.10} & \textbf{15.53} & & \textbf{43.13} & \textbf{11.64} & \textbf{17.35}\\
\bottomrule
\end{tabular}}
\caption{Automatic evaluation results of each summary segment and combined summary on the BigSurvey-MDS test set.}
\label{table:autoeval_bigsurveymds}
\end{table*}

To deal with the above problem, we use category-based alignment (CA) to align each summary section with input sentences classified as the same type. The aligned input text and target summary compose the example for model training.
CA can be regarded as a content selection operation based on sentence classification, supporting the summarization model to focus on specific aspects of input documents.

Considering that different sections of the structured summary can be written in different ways, we train multiple models producing separate sections in the target summary.
To prepare the pairs of input and output for model training, aligning all summary sections with the same input (one-to-many) is straightforward. 
Merged from dozens of reference papers' abstracts, the input text of each example in the BigSurvey-MDS usually contains multiple aspects of content. For a summary section focusing on a specific aspect, other aspects of input content can be regarded as noises. Using the same input for producing different summary sections can make the produced sections mix different aspects of content.

Classifying sentences from reference papers' abstracts can be defined as a sequential sentence classification (SSC) problem. We follow the setting in \cite{cohan2019pretrained} and train a BERT-based SSC model on the datasets named CSABST \cite{cohan2019pretrained}. 
We first use the SSC model to classify the sentences in each reference paper's abstract and then merge the sentences classified as the same type.
Our evaluation results show that the SSC model can outperform the SCC model in the abstract sentences classification. 
Considering the target summary in BigSurvey-MDS is usually much longer than the samples in the CSABST and the max length limit of the BERT model, it is not appropriate to use the SSC model trained on the CSABST to classify the target summary's sentences. Therefore, we utilize the SSC model to classify input sentences and the SCC model to classify sentences in the target summary.

The original transformer model's encoder adopts the self-attention mechanism scaling quadratically with the number of tokens in input sequences \cite{vaswani2017attention}. It is prohibitively expensive for the long input sequence \cite{choromanski2020rethinking} and precludes fine-tuning large pre-trained models with limited computational resources. Some transformer models' variants adopt sparse attention mechanisms to reduce the complexity. For example, BigBird \cite{zaheer2020big} and Longformer \cite{beltagy2020longformer} combine three different types of attention mechanisms and scale linearly with sequence length. 
Considering the constraint of GPU memory, our CAST model employs the pre-trained encoder with sparse attention to encode longer input texts.
Our CAST model has two versions, the CAST-BigBird employs the BigBird \cite{zaheer2020big} as the encoder, and the CAST-LED's encoder is from the Longformer \cite{beltagy2020longformer}.

\section{Experiments}
\label{sec:experiments}

\subsection{Baselines}
In our experiments, we compare various extractive and abstractive summarization models on our BigSurvey dataset. 

\paragraph{LexRank and TextRank.}Two unsupervised extractive summarizers are built on graph-based ranking methods \cite{erkan2004lexrank,mihalcea2004textrank}.

\paragraph{CopyTransformer.}Gehrmann et al.~\shortcite{gehrmann2018bottom} add the copy mechanism \cite{see2017get} to the transformer model for abstractive summarization.

\paragraph{BART.}Lewis et al.~\shortcite{lewis2020bart} build a sequence-to-sequence denoising autoencoder that is pre-trained to reconstruct the original input text from the corrupted text. 

\paragraph{PEGASUS.}Zhang et al.~\shortcite{zhang2020pegasus} pre-train a transformer-based model with the Gap Sentences Generation (GSG) and Masked Language Model (MLM) objectives.

\paragraph{BigBird-PEGASUS.}Zaheer et al.~\shortcite{zaheer2020big} combine the BigBird encoder with the decoder from the PEGASUS model.

\paragraph{Longformer-Encoder-Decoder (LED).}LED \cite{beltagy2020longformer} is built on BART and adopts the local and global attention mechanisms in the encoder part, while its decoder part still utilizes the original self-attention mechanism.

\noindent We fine-tuned large models of these pre-trained summarizers on BigSurvey's training set.

\subsection{Experimental Setting}

The vocabulary's maximum size is set as 50,265 for these abstractive summarization models, while the BERT-based classifiers use 30,522 as default. We use dropout with the probability 0.1. The optimizer is Adam with $\beta_1$=0.9 and $\beta_2$=0.999. Summarization models use learning rate of $5e^{-5}$, while the classifiers use $2e^{-5}$. We also adopt the learning rate warmup and decay. During decoding, we use beam search with a beam size of 5. Trigram blocking is used to reduce repetitions. 
We adopt the implementations of PEGASUS, BigBird, and LED from HuggingFace's Transformers \cite{wolf2020transformers}. The BART's implementation is from the fairseq \cite{ott2019fairseq}. All the models are trained on one NVIDIA RTX8000.

\begin{table}[t]
\renewcommand\arraystretch{1.0}
\small
\centering
\begin{tabular}{lccc}
\toprule
\textbf{Method} & \textbf{R-1} & \textbf{R-2} & \textbf{R-L}\\
\midrule
LexRank & 30.93 & 8.53 & 15.54 \\
TextRank & 32.21 & 8.79 & 15.96 \\
\midrule
CopyTransformer & 30.59 & 5.80 & 16.76 \\
BART & 35.28 & 9.71 & 17.89 \\
PEGASUS & 37.47 & 11.08 & 19.25 \\
LED & 38.57 & 11.52 & 19.36 \\
BigBird-PEGASUS & \textbf{39.75} & \textbf{12.60} & \textbf{20.11} \\
\bottomrule
\end{tabular}
\caption{Automatic evaluation results on the BigSurvey-Abs.}
\label{table:autoeval_bigsurveyabs}
\end{table}

\subsection{Results and Discussion}
In our experiments, we train and evaluate various summarization models on the BigSurvey-MDS and BigSurvey-Abs. 
We divide the BigSurvey-MDS into three subsets and train three models producing separate sections in the target summary. In this section, we report and analyze our experimental results. 

To compare the quality of summaries produced by these models, we conduct automatic evaluation and report the ROUGE $\mathrm{F}_1$ scores \cite{lin2004rouge}, including the overlap of unigrams (R-1), bigrams (R-2), and longest common subsequence (R-L). 
We report ROUGE scores of produced three summary sections and the combined summaries for the BigSurvey-MDS in Table \ref{table:autoeval_bigsurveymds}. It shows that these abstractive summarization models can outperform these extractive models on the BigSurvey-MDS.
Replacing the encoder's self-attention mechanism with sparse attention mechanisms can enable us to train transformer-based models on longer input texts with limited GPU memory. The BigBird-PEGASUS and the LED outperform other transformer-based models without sparse attention, which reveals that introducing longer input text can benefit the quality of generated summaries.

The concatenation of input reference papers' abstracts usually contains multiple aspects of content. It requires the summarization method to have a strong capability of content selection to produce a summary section precisely covering a specified aspect. 
We compare the effects of applying different ways of alignment (one-to-many or category-based alignment) on various summarization models.
When using the one-to-many alignment, we observe that the produced summary sections often mix multiple aspects of content and have overlapping content in different sections. It reveals that these summarization models still have difficulties in content selection, although they have supervision from target summaries. 
Table \ref{table:autoeval_bigsurveymds} shows that introducing CA can bring extra performance gains for various summarization models. It reflects the effectiveness of CA and the need to enhance summarization models' capabilities of content selection.
Combing the CA and the transformer model with sparse attention mechanism, CAST-LED outperforms other baseline models on the BigSurvey-MDS.

Table \ref{table:autoeval_bigsurveyabs} shows the evaluation results on the BigSurvey-Abs. These transformer-based abstractive summarization models with sparse attention mechanisms also outperform other baselines. It reveals that modeling longer input text is also important for summarizing survey papers in the BigSurvey-Abs. 
Besides, the pre-trained sequence-to-sequence models outperform the model trained from scratch.

In addition to automatic evaluation, we performed a human evaluation to compare two summarization models' generated summaries in terms of their informativeness (the coverage of input documents' content), fluency (content organization and grammatical correctness), and non-redundancy (fewer repetitions). 
We randomly selected 50 samples from the test set of the BigSurvey-MDS. Four annotators are required to compare two models' generated summaries that are presented anonymously. We also assess their agreements by Fleiss' kappa \cite{fleiss1971measuring}. Human evaluation results in Table \ref{table:humaneval_results} exhibit that our CAST-LED method outperforms the original LED model in terms of informativeness and non-redundancy.

\begin{table}[t]
\small
\renewcommand\arraystretch{1.0}
\centering
\begin{tabular}{lcccc}
\toprule & \textbf{Win} & \textbf{Lose} & \textbf{Tie} & \textbf{Kappa}\\
\midrule
Informativeness & 39.5\% & 25.0\%& 35.5\%& 0.659\\
Fluency & 28.5\%& 27.5\%& 44.0\%& 0.631\\
Non-Redundancy & 33.0\%& 25.5\%& 41.5\%& 0.623\\
\bottomrule
\end{tabular}
\caption{Human evaluation results on the test set of BigSurvey-MDS. "Win" denotes that the generated summary of our CAST-LED is better than that of the original LED model in one aspect. "Tie" represents that two summaries are comparable in one aspect.}
\label{table:humaneval_results}
\end{table}

\begin{table}[t]
\small
\renewcommand\arraystretch{1.0}
\centering
\begin{tabular}{lccc}
\toprule
 & \textbf{R-1} & \textbf{R-2} & \textbf{R-L}\\
\midrule
CAST-LED & \textbf{43.13} & \textbf{11.64} & \textbf{17.35}\\
w/o sparse attn & 40.21 & 9.38 & 16.06 \\
w/o CA & 39.79 & 9.42 & 16.05\\
w/o CA + $\mathrm{LED}_{base\mbox{-}8192}$ & 39.38 & 9.78 & 16.30\\
\bottomrule
\end{tabular}
\caption{\label{table:evalablation} Ablation study on the test set of BigSurvey-MDS. We report the ROUGE scores of combined summaries. "w/o sparse attn" denotes using the original self-attention in the encoder. "w/o CA" represents removing the category-based alignment.} 
\end{table}

We also conduct the ablation study to validate the effectiveness of individual components in our method. Table \ref{table:evalablation} shows that using the original self-attention to replace the sparse attention mechanism in the encoder part or removing the category-based alignment can lead to performance degradation. Besides, increasing the input sequence length cannot replace the CA. The longer inputs can introduce more noises, and it is still difficult for summarization models to select the salient content on the specific aspect without CA. The results verify the effectiveness of the sparse attention mechanism and CA. 

\section{Conclusion}
In this paper, we introduce BigSurvey, the first large-scale dataset for numerous academic papers summarization. It is built on human-written survey papers and their reference papers. BigSurvey includes two subsets for producing two-level summaries. Besides, we propose a method named category-based alignment and sparse transformer (CAST) to generate the structured summary covering dozens of papers on a research topic. Dataset analyses and experimental results reveal the importance of adopting the category-based alignment and sparse attention mechanism. 
When observing generated summaries, we find that these summarization models still lack criticism and reasoning abilities, and their generated summaries are not yet comparable to human-written summaries.


\section*{Acknowledgments}

This work described in this paper was supported by grants from the Collaborative Research Fund sponsored by the Research Grants Council of HKSAR, China (Project No.C5026-18G and C6030-18G) and the Hong Kong Jockey Club Charities Trust (Project S/N Ref.: 2021-0369).

\bibliographystyle{named}
\bibliography{ijcai22}

\begin{thebibliography}{}

\bibitem[\protect\citeauthoryear{Ammar \bgroup \em et al.\egroup
  }{2018}]{ammar2018construction}
Waleed Ammar, Dirk Groeneveld, et~al.
\newblock Construction of the literature graph in semantic scholar.
\newblock In {\em NAACL-HLT}, 2018.

\bibitem[\protect\citeauthoryear{Aviv-Reuven and
  Rosenfeld}{2021}]{aviv2021publication}
Shir Aviv-Reuven and Ariel Rosenfeld.
\newblock Publication patterns’ changes due to the covid-19 pandemic: A
  longitudinal and short-term scientometric analysis.
\newblock {\em Scientometrics}, pages 1--24, 2021.

\bibitem[\protect\citeauthoryear{Beltagy \bgroup \em et al.\egroup
  }{2020}]{beltagy2020longformer}
Iz~Beltagy, Matthew~E Peters, and Arman Cohan.
\newblock Longformer: The long-document transformer.
\newblock {\em arXiv preprint arXiv:2004.05150}, 2020.

\bibitem[\protect\citeauthoryear{Bird \bgroup \em et al.\egroup
  }{2009}]{bird2009natural}
Steven Bird, Ewan Klein, and Edward Loper.
\newblock {\em Natural language processing with Python}.
\newblock O'Reilly Media, Inc., 2009.

\bibitem[\protect\citeauthoryear{Choromanski \bgroup \em et al.\egroup
  }{2020}]{choromanski2020rethinking}
Krzysztof Choromanski, Valerii Likhosherstov, et~al.
\newblock Rethinking attention with performers.
\newblock {\em arXiv preprint arXiv:2009.14794}, 2020.

\bibitem[\protect\citeauthoryear{Cohan \bgroup \em et al.\egroup
  }{2019}]{cohan2019pretrained}
Arman Cohan, Iz~Beltagy, King, et~al.
\newblock Pretrained language models for sequential sentence classification.
\newblock In {\em EMNLP-IJCNLP}, pages 3693--3699, 2019.

\bibitem[\protect\citeauthoryear{Devlin and others}{2019}]{devlin2019bert}
Jacob Devlin et~al.
\newblock Bert: Pre-training of deep bidirectional transformers for language
  understanding.
\newblock In {\em NAACL-HLT}, pages 4171--4186, 2019.

\bibitem[\protect\citeauthoryear{Erkan and Radev}{2004}]{erkan2004lexrank}
G{\"u}nes Erkan and Dragomir~R Radev.
\newblock Lexrank: Graph-based lexical centrality as salience in text
  summarization.
\newblock {\em Journal of artificial intelligence research}, pages 457--479,
  2004.

\bibitem[\protect\citeauthoryear{Fabbri \bgroup \em et al.\egroup
  }{2019}]{fabbri2019multi}
Alexander~Richard Fabbri, Irene Li, Tianwei She, et~al.
\newblock Multi-news: A large-scale multi-document summarization dataset and
  abstractive hierarchical model.
\newblock In {\em ACL}, pages 1074--1084, 2019.

\bibitem[\protect\citeauthoryear{Fleiss}{1971}]{fleiss1971measuring}
Joseph~L Fleiss.
\newblock Measuring nominal scale agreement among many raters.
\newblock {\em Psychological bulletin}, page 378, 1971.

\bibitem[\protect\citeauthoryear{Gehrmann \bgroup \em et al.\egroup
  }{2018}]{gehrmann2018bottom}
Sebastian Gehrmann, Yuntian Deng, and Alexander Rush.
\newblock Bottom-up abstractive summarization.
\newblock In {\em EMNLP}, pages 4098--4109, 2018.

\bibitem[\protect\citeauthoryear{Gidiotis and Tsoumakas}{2019}]{GidiotisT19}
Alexios Gidiotis and Grigorios Tsoumakas.
\newblock Structured summarization of academic publications.
\newblock In {\em PKDD/ECML}, pages 636--645, 2019.

\bibitem[\protect\citeauthoryear{Grusky \bgroup \em et al.\egroup
  }{2018}]{grusky-etal-2018-newsroom}
Max Grusky, Mor Naaman, and Yoav Artzi.
\newblock {N}ewsroom: A dataset of 1.3 million summaries with diverse
  extractive strategies.
\newblock In {\em NAACL-HLT}, pages 708--719, 2018.

\bibitem[\protect\citeauthoryear{Hartley}{2004}]{hartley2004current}
James Hartley.
\newblock Current findings from research on structured abstracts.
\newblock {\em Journal of the Medical Library Association}, page 368, 2004.

\bibitem[\protect\citeauthoryear{Hartley}{2014}]{hartley2014current}
James Hartley.
\newblock Current findings from research on structured abstracts: An update.
\newblock {\em Journal of the Medical Library Association}, pages 146--148,
  2014.

\bibitem[\protect\citeauthoryear{Lewis \bgroup \em et al.\egroup
  }{2020}]{lewis2020bart}
Mike Lewis, Yinhan Liu, Naman Goyal, et~al.
\newblock Bart: Denoising sequence-to-sequence pre-training for natural
  language generation, translation, and comprehension.
\newblock In {\em ACL}, pages 7871--7880, 2020.

\bibitem[\protect\citeauthoryear{Lin}{2004}]{lin2004rouge}
Chin-Yew Lin.
\newblock Rouge: A package for automatic evaluation of summaries.
\newblock In {\em Text summarization branches out}, pages 74--81, 2004.

\bibitem[\protect\citeauthoryear{Liu \bgroup \em et al.\egroup
  }{2018}]{liu2018generating}
Peter~J Liu, Mohammad Saleh, Etienne Pot, et~al.
\newblock Generating wikipedia by summarizing long sequences.
\newblock In {\em ICLR}, 2018.

\bibitem[\protect\citeauthoryear{Liu \bgroup \em et al.\egroup
  }{2019}]{liu2019roberta}
Yinhan Liu, Myle Ott, Naman Goyal, et~al.
\newblock Roberta: A robustly optimized bert pretraining approach.
\newblock {\em arXiv preprint arXiv:1907.11692}, 2019.

\bibitem[\protect\citeauthoryear{Liu \bgroup \em et al.\egroup
  }{2021}]{liu2021highlight}
Shuaiqi Liu, Jiannong Cao, et~al.
\newblock Highlight-transformer: Leveraging key phrase aware attention to
  improve abstractive multi-document summarization.
\newblock In {\em Findings of the ACL-IJCNLP}, pages 5021--5027, 2021.

\bibitem[\protect\citeauthoryear{Liu \bgroup \em et al.\egroup
  }{2022}]{LIU2022102913}
Shuaiqi Liu, Jiannong Cao, et~al.
\newblock Key phrase aware transformer for abstractive summarization.
\newblock {\em Information Processing \& Management}, 59(3):102913, 2022.

\bibitem[\protect\citeauthoryear{Lu \bgroup \em et al.\egroup
  }{2020}]{lu-etal-2020-multi-xscience}
Yao Lu, Yue Dong, and Laurent Charlin.
\newblock Multi-{XS}cience: A large-scale dataset for extreme multi-document
  summarization of scientific articles.
\newblock In {\em EMNLP}, pages 8068--8074, 2020.

\bibitem[\protect\citeauthoryear{Meng \bgroup \em et al.\egroup
  }{2021}]{meng2021bringing}
Rui Meng, Khushboo Thaker, Lei Zhang, et~al.
\newblock Bringing structure into summaries: a faceted summarization dataset
  for long scientific documents.
\newblock In {\em ACL-IJCNLP}, pages 1080--1089, 2021.

\bibitem[\protect\citeauthoryear{Mihalcea and
  Tarau}{2004}]{mihalcea2004textrank}
Rada Mihalcea and Paul Tarau.
\newblock Textrank: Bringing order into text.
\newblock In {\em EMNLP}, pages 404--411, 2004.

\bibitem[\protect\citeauthoryear{Ott and others}{2019}]{ott2019fairseq}
Myle Ott et~al.
\newblock fairseq: A fast, extensible toolkit for sequence modeling.
\newblock In {\em NAACL-HLT}, 2019.

\bibitem[\protect\citeauthoryear{See \bgroup \em et al.\egroup
  }{2017}]{see2017get}
Abigail See, Peter~J. Liu, and Christopher~D. Manning.
\newblock Get to the point: Summarization with pointer-generator networks.
\newblock In {\em ACL}, pages 1073--1083, 2017.

\bibitem[\protect\citeauthoryear{Sinha \bgroup \em et al.\egroup
  }{2015}]{sinha2015overview}
Arnab Sinha, Zhihong Shen, Yang Song, et~al.
\newblock An overview of microsoft academic service (mas) and applications.
\newblock In {\em WWW}, pages 243--246, 2015.

\bibitem[\protect\citeauthoryear{Vaswani and
  others}{2017}]{vaswani2017attention}
Ashish Vaswani et~al.
\newblock Attention is all you need.
\newblock In {\em NeurIPS}, pages 5998--6008, 2017.

\bibitem[\protect\citeauthoryear{Wolf \bgroup \em et al.\egroup
  }{2020}]{wolf2020transformers}
Thomas Wolf, Julien Chaumond, Lysandre Debut, et~al.
\newblock Transformers: State-of-the-art natural language processing.
\newblock In {\em EMNLP}, pages 38--45, 2020.

\bibitem[\protect\citeauthoryear{Zaheer \bgroup \em et al.\egroup
  }{2020}]{zaheer2020big}
Manzil Zaheer, Guru Guruganesh, Kumar~Avinava Dubey, et~al.
\newblock Big bird: Transformers for longer sequences.
\newblock In {\em NeurIPS}, 2020.

\bibitem[\protect\citeauthoryear{Zhang and others}{2020}]{zhang2020pegasus}
Jingqing Zhang et~al.
\newblock Pegasus: Pre-training with extracted gap-sentences for abstractive
  summarization.
\newblock In {\em ICML}, pages 11328--11339, 2020.

\end{thebibliography}

\end{document}